# The Expertise Level

Ron Fulbright

University of South Carolina Upstate
800 University Way, Spartanburg, SC 29303
`rfulbright@uscupstate.edu`

**Abstract.** Computers are quickly gaining on us. Artificial systems are now exceeding the performance of human experts in several domains. However, we do not yet have a deep definition of expertise. This paper examines the nature of expertise and presents an abstract knowledge-level and skill-level description of expertise. A new level lying above the Knowledge Level, called the Expertise Level, is introduced to describe the skills of an expert without having to worry about details of the knowledge required. The Model of Expertise is introduced combining the knowledge-level and expertise-level descriptions. Application of the model to the fields of cognitive architectures and human cognitive augmentation is demonstrated and several famous intelligent systems are analyzed with the model.

## 1      Introduction

Artificial systems are gaining on us! Powered by new machine learning and reasoning methods, artificial systems are beginning to exceed expert human performance in many domains. IBM's Deep Blue, defeated the reigning human chess champion in 1997 [1]. In 2011, a cognitive system built by IBM, called Watson, defeated the two most successful human champions of all time in the game of Jeopardy! [2, 3]. In 2016, Google's AlphaGo defeated the reigning world champion in Go, a game vastly more complex than Chess [4, 5]. In 2017, a version called AlphaGo Zero learned how to play Go by playing games with itself not relying on any data from human games [6]. AlphaGo Zero exceeded the capabilities of AlphaGo in only three days. Also in 2017, a generalized version of the learning algorithm called AlphaZero was developed capable of learning any game. After only a few hours of self-training, AlphaZero achieved expert-level performance in the games of Chess, Go, and Shogi [7].

This technology goes far beyond playing games. Computers are now better at predicting mortality than human doctors [8], detecting early signs of heart failure [9], detecting signs of child depression through speech [10], and can even find discoveries in old scientific papers missed by humans [11]. Many other examples of artificial systems achieving expert-level performance exist.

## 2      Literature

### 2.1 What is an expert?
What does it mean to be an expert? What is expertise? Are these systems really artificial experts? To answer these kinds of questions, one needs a model of expertise to compare them to. The nature of intelligence and expertise has been debated for decades. To motivate the Model of Expertise presented in this paper, we draw from research in artificial intelligence, cognitive science, intelligent agents, and educational pedagogy.

As Gobet points out, traditional definitions of expertise rely on knowledge (what an expert knows) and skills (what an expert knows how to do) [12]. Some definitions say an expert knows more than a novice while other definitions say an expert can do more than a novice. While an expert is certainly expected to know about their topic and be able to perform skills related to that topic, not everyone who is knowledgeable and skillful in a domain is an expert in that domain. Simply knowing more and being able to do more is not enough. Therefore, Gobet gives a results-based definition of expertise:

> "…an expert is somebody who obtains results vastly superior to those obtained by the majority of the population."

This definition immediately runs into the venerable debates involving Searle's Chinese room [13] and the Turing test [14]. Is a machine yielding results like an expert really an expert? Answering these kinds of questions is difficult because we lack a deep model of expertise. In their influential study of experts, Chase and Simon state: [15]

> "…a major component of expertise is the ability to recognize a very large number of specific relevant cues when they are present in any situation, and then to retrieve from memory information about what to do when those particular cues are noticed. Because of this knowledge and recognition capability, experts can respond to new situations very rapidly and usually with considerable accuracy."

Experts look at a current situation and match it to an enormous store of domain-specific knowledge. Steels later describes this as deep domain knowledge [16]. Experts acquire this enormous amount of domain knowledge from experience (something we now call episodic memory). It is estimated experts possesses at least 50,000 pieces of domain-specific knowledge requiring on the order of 10,000 hours of experience. Even though these estimates have been debated, it is generally agreed experts possess vast domain knowledge and experience. Experts extract from memory much more knowledge, both implicit and explicit, than novices. Furthermore, an expert applies this greater knowledge to the situation at hand more efficiently and quickly than a novice. Therefore, experts are better and more efficient problem solvers than novices.

The ability of an expert to quickly jump to the correct solution has been called intuition. A great deal of effort has gone into defining and studying intuition. Dreyfus and Dreyfus (Dreyfus, 1972) and (Dreyfus & Dreyfus, 1988) argue intuition is a holistic human property not able to be captured by a computer program [17, 18]. However, Simon et al. argue intuition is just the ultra-efficient matching and retrieval of "chunks" of knowledge and know-how.

The idea of a "chunk" of information has been associated with artificial intelligence research and cognitive science for decades dating back to pioneers Newell and Simon. Gobet and Chassy argue the traditional notion of a "chunk" is too simple and instead, introduce the notion of a "template" as a chunk with static components and variable, or dynamic components, resembling a complex data structure [19]. Templates are similar to other knowledge representation mechanisms in artificial intelligence such as Minsky frames [20] and models in intelligent agent theory [21]. As Gobet and Simon contend, templates allow an expert to quickly process information at different levels of abstraction yielding the extreme performance consistent with intuition [22].

DeGroot experimentally established the importance of perception in expertise [23]. When perceiving a situation in the environment, an expert is able to see the most important things quicker than a novice. Being able to perceive the most important cues and retrieve knowledge form one's experience and quickly apply it are hallmarks of expertise.

Steels identified the following as needed by experts: deep domain knowledge, problem-solving methods, and task models [16]. Problem-solving methods are how one goes about solving a problem. There are generic problem solutions applicable to almost every domain of discourse such as "how to average a list of numbers." However, there are also domain-specific problem-solving methods applicable to only a specific domain or very small collection of domains or even just one domain. A task model is knowledge about how to do something. For example, how to remove a faucet is a task an expert plumber would know. As with problem solutions, there are generic tasks and domain-specific tasks. Summarizing, the basic requirements for an expert are:

- the ability to experience and learn domain knowledge
- learn task knowledge
- learn problem-solving knowledge
- perceive a current situation
- match the current situation with known domain knowledge
- retrieve knowledge relevant to the situation
- apply knowledge and know-how
- achieve superior results

**2.2 The Knowledge Level**
Cognitive scientists have studied human cognition for many decades with the hope of being able to create artificial entities able to perform human-level cognition. Newell recognized computer systems are described at many different levels and defined the Knowledge Level as a means to analyze the knowledge of intelligent agents at an abstract level as shown in Fig. 1 [24]. The lower levels represent physical elements from which the system is constructed (electronic devices and circuits). The higher levels represent the logical elements of the system (logic circuits, registers, symbols in programs).

In general, a level "abstracts away" the details of the lower level. For example, consider a computer programmer writing a line of code storing a value into a variable. The programmer is operating at the Program/Symbol Level and never thinks about how the value is stored in registers and ultimately is physically realized as voltage potentials in electronic circuits. Details of the physical levels are abstracted away at the Program Level. Likewise, at the Knowledge Level, implementation details of how knowledge is represented in computer programs is abstracted away. This allows us to talk about knowledge in implementation-independent terms facilitating generic analysis about intelligent agents.

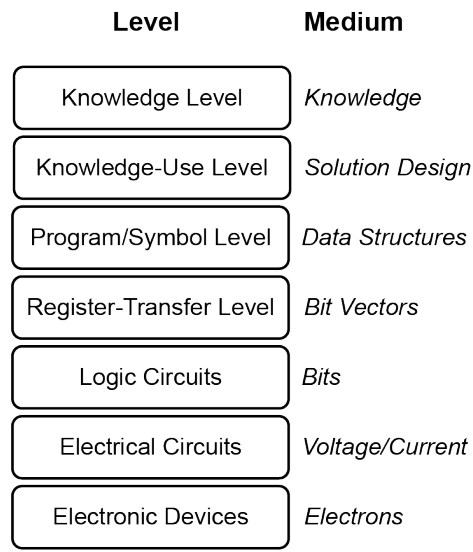

**Fig. 1.** The Knowledge Level

Steels added the Knowledge-Use Level between the Knowledge Level and the Program Level to address issues like task decomposition, execution, scheduling, software architecture, and data structure design [16]. This level is geared toward implementation and is quite dependent on implementation details but is necessary to bridge the gap between the Knowledge Level and the Program/Symbol Level.

**2.3 Cognitive Architecture**

Cognitive scientists have spent much effort analyzing and modeling human intelligence and cognition. One of the most successful models of human cognition is the Soar model shown in Fig. 2 began by Newell and evolved by his students and others for over thirty years [25]. Although not explicitly a part of the Soar model, the figure shows the Soar model situated in an environment with *perceive* and *act* functions. The agent can learn new procedural knowledge (how to do things), semantic knowledge (knowledge about things), and episodic knowledge (knowledge about its experiences). New knowledge can be learned by reinforcement learning, semantic learning, or episodic learning.

Pieces of this knowledge from long term memory, as well as perceptions, are brought into a working area of memory, short term memory, where they are processed by the appraisal and decision functions. Soar is a model of human cognition but is not necessarily a model of expertise. We will later update the Soar architecture to include elements to support expertise.

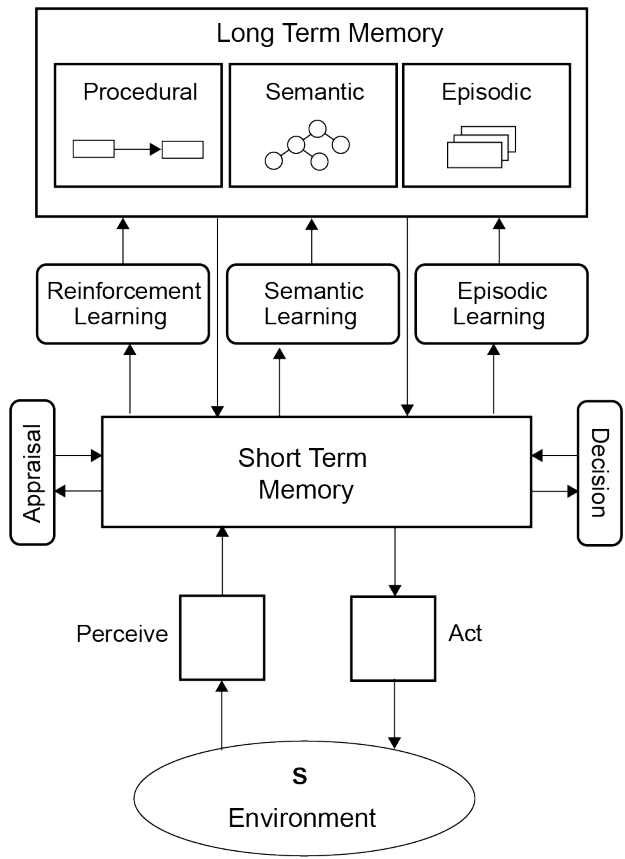

**Fig. 2.** The Soar model of cognition

### 2.4 Formal Intelligent Agent Models

Researchers in artificial intelligence have defined several architectural models of intelligent agents. These models are of interest to us in this paper because an expert is certainly an intelligent agent. Fig. 3 shows a formal model of a goal-driven, utility-based learning/evolving intelligent agent [21, 26].

Situated in an environment, intelligent agents repeatedly execute the perceive-reason-act-learn cycle. Through various sensors, the *see* function allows agents to perceive the environment, *S*, as best they can. Agents can perceive only a subset, *T*, of the environment. Every agent has a set of actions, *A*, they can perform via the *do* function. The agent selects the appropriate action through the *action* function. Every action causes the environment to change state.

Models, *M*, are internal descriptions of objects, situations, and the real world. The *model* function matches incomplete data from the agent's perceptions with its models to classify what it is currently encountering. For example, a "danger" model would allow an agent to recognize a hazardous situation.

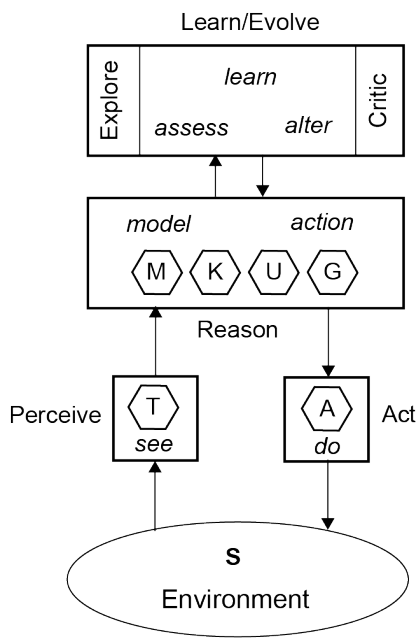

{*K, M, U, G, S, T, A,* see, do, action, model, learn, alter, assess}

| | |
|---|---|
| *K* | Set of general knowledge statements |
| *M* | Set of internal states/models |
| *U* | Set of utility values |
| *G* | Set of goals the agent is trying to achieve |
| *S* | Set of states for the environment |
| *T* | Partitions of *S* distinguishable by the agent |
| *A* | Set of actions the agent can perform |
| see | $S \rightarrow T$ |
| do | $A \times S \rightarrow S$ |
| action | $M \times G \times K \times U \times T \rightarrow A$ |
| model | $M \times G \times K \times U \times T \rightarrow M$ |
| learn | $M \times G \times K \times U \times T \rightarrow K$ |
| alter | $M \times G \times K \times U \times T \rightarrow G$ |
| assess | $M \times G \times K \times U \times T \rightarrow U$ |

**Fig. 3.** Formal model of a learning/evolving intelligent agent.

Intelligent agents continually work to achieve one or more goals, **G**. The *alter* function allows the agent to change its goals over time. Agents can change their sensitivity to things using a set of utility values, **U**. The *assess* function allows the agent to adjust its utility model. This is important because the urgency of certain actions and goals changes over time and with changes in a dynamic environment. For example, the goal "recharge batteries" might have a low utility value at the beginning of a journey but as the battery charge level gets lower, the utility value of the "recharge batteries" goal rises and eventually becomes the most important goal. An intelligent agent maintains a set of general knowledge, **K**, and can learn new knowledge.

### 2.5 Bloom's Taxonomy

There are other notions of "expertise" from outside the fields of artificial intelligence and cognitive science. Originally published in the 1950s, and revised in the 1990s, Bloom's Taxonomy was developed in the educational field as a common language about learning and educational goals to aid in the development of courses and assessments [27, 28].

As shown in Fig. 4, Bloom's taxonomy consists of a set of verbs describing fundamental cognitive processes at different levels of difficulty. The idea behind Bloom's Taxonomy is when learning a new subject, a student able to perform these processes, has demonstrated proficiency in the subject matter. The processes are listed in order from the simplest (remember) to the most complex (create). Here, we propose these processes also describe expert performance in a particular domain. The verbs in Bloom's Taxonomy identify skills we expect an expert to possess.

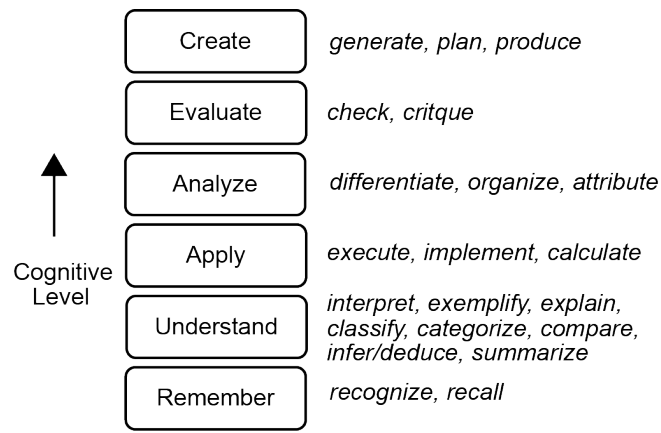

**Fig. 4.** Bloom's Taxonomy

We note two things as we look at historical notions of intelligence and expertise. First, we encounter both *knowledge* and *skills* as requirements of an expert. Therefore, a model of expertise must be able to represent both. Second, while there are many models of intelligence and human cognition, there are no comprehensive models of expertise. While it may be true all experts are intelligent agents, it is not true all intelligent agents are experts. Therefore, we seek a comprehensive model of expertise and wish for this model to be implementation independent. Our model should apply to human and artificial experts.

In this paper, we first introduce a new abstract level called the *Expertise Level*. Lying above the Knowledge Level, the Expertise Level describes the *skills* needed by an expert. We use the skills identified by Simon, Steels, and Gobet augmented with the skills from Bloom's Taxonomy to form this description. We also develop a Knowledge Level description of expertise describing the knowledge required of an expert. For this description, we combine the knowledge identified in various cognitive architectures discussed earlier. We then show how to apply the Model of Expertise by showing how it can be incorporated into the Soar architecture and how it can be used in the field of human cognitive augmentation. We then use the Model of Expertise to discuss and characterize current systems.

## 3 The Model of Expertise

### 3.1 The Expertise Level

As described earlier, Simon, Steels, and Gobet identify kinds of knowledge an expert must have and kinds of functions or actions an expert must perform. Newell's Knowledge Level is suitable for holding a description of an expert's knowledge. However, a full model of expertise must accommodate both knowledge and skills.

As shown in Fig. 5, we extend Newell's levels and create a new level called the *Expertise Level* above the Knowledge Level to represent skills an expert must possess. At the Expertise Level, we talk about what an expert does—the skills—and not worry about the details of the knowledge required to perform these skills. Therefore, the medium of the Expertise Level is *skills*.

| Level | Medium |
|---|---|
| Expertise Level | *Skills* |
| Knowledge Level | *Knowledge* |
| Knowledge-Use Level | *Solution Design* |
| Program/Symbol Level | *Data Structures* |
| Register-Transfer Level | *Bit Vectors* |
| Logic Circuits | *Bits* |
| Electrical Circuits | *Voltage/Current* |
| Electronic Devices | *Electrons* |

**Fig. 5.** The Expertise Level

### 3.2 The Expertise Level Description of Experts

What skills does an expert need to have? We start with the six skills identified in Bloom's Taxonomy: *recall*, *understand*, *apply*, *analyze*, *evaluate*, and *create*. These skills were identified because in the education field, a student demonstrating ability in all six skills is considered to have achieved a mastery of the subject matter. An expert certainly has mastery of subject matter, so any model of expertise should include these six skills.

An expert certainly is considered to be an intelligent agent operating in an environment. As such, following the example set by researchers from the intelligent agent theory field, the expert must sense the environment (*perceive*) and perform actions (*act*) to effect changes on the environment. Common in the intelligent agent and cognitive science fields is the notion of learning. An expert acquires knowledge and know-how via experience through the *learn* skill. Future refinement of the model may identify several different kinds of learning and therefore add skills, but here we represent all types of learning with the single *learn* skill. From the work of Simon, Steels, and Gobet, the *extract* skill represents the expert's ability to match perceptions to stored deep domain knowledge and knowledge, procedures, and tasks relevant to the situation.

Experts are goal-driven intelligent agents with the ability to change goals over time. The *alter* skill allows the expert to change its goals as it evolves. We also believe experts must be utility-driven intelligent agents and must have the ability to adjust its utility values. We do not create a new skill for this because this ability is subsumed by the *evaluate* skill already in our list of skills. We feel an expert should be able to *teach* about their domain of discourse, so include this skill in our list.

Therefore, as shown in Fig. 6, twelve skills are identified at the Expertise Level description of an expert: *recall, apply, evaluate, understand, analyze, create, extract, teach, perceive, learn, alter, and act.*

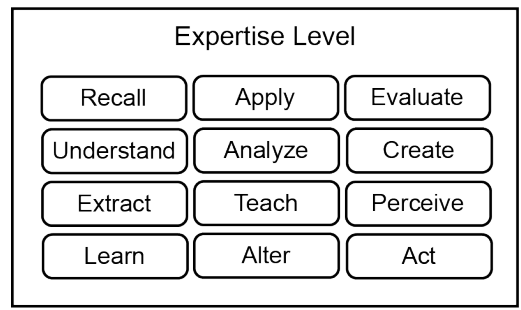

**Fig. 6.** The Expertise Level Description of Experts

### 3.3 The Knowledge Level Description of Experts

To create the Knowledge Level description of an expert, we combine ideas from intelligent agent theory, cognitive architectures, and cognitive science. An expert is an agent able to perceive the environment. Because of limitations in its sensory systems, an expert perceives only a subset of the possible states of the environment, $T$. The expert also has a set of actions, $A$, it can perform to change the environment. Because experts are goal-driven and utility-driven evolving agents, $G$ represents the set of goals and $U$ represents the set of utility values.

In addition to deep domain knowledge $K_D$ (knowledge about the domain) experts possess general background knowledge $K$ (generic knowledge about things), common-sense knowledge $K_C$, and episodic knowledge $K_E$ (knowledge from and about experiences). A model, similar to Gobet's templates and Minksy's frames, is an internal representation allowing the expert to classify its perceptions and recognize or differentiate situations it encounters. For example, an expert plumber would have an idea of what a leaky faucet looks, sounds, and acts like based on experience allowing the plumber to quickly recognize a leaky faucet. Some models are domain-specific, $M_D$, and other models are generic, $M$. In humans, the collection and depth of models is attained from years of experience. As models are learned from experience, creating and maintaining $M_D$ requires $K_E$ and $K_D$ as a minimum but may also involve other knowledge stores. Following Simon and Steels, an expert must know how to solve problems in a generic sense, $P$, and how to solve problems with domain-specific methods, $P_D$. In addition, experts must also know how to perform generic tasks, $L$, and domain-specific tasks, $L_D$.

Therefore, we have identified 14 knowledge stores an expert maintains as shown in Fig. 7.

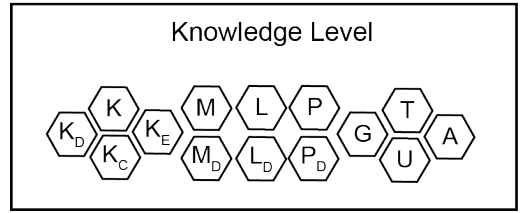

**Fig. 7.** The Knowledge Level Description of Experts

### 3.4 The Model of Expertise
Combining the Expertise Level description of experts shown in Fig. 6 with the Knowledge Level description shown in Fig. 7 yields the full Model of Expertise shown in Fig. 8. Experts require 12 fundamental skills: *recall, apply, evaluate, understand, analyze, create, extract, teach, perceive, learn, alter,* and *act*. When working with another entity, experts need the *collaborate* skill as well.

Experts also require the maintenance of 14 knowledge stores: 4 general types of knowledge (generic, domain-specific, common-sense, and episodic), generic and domain-specific models, generic and domain-specific task models, generic and domain-specific problem-solving models, a set of goals, utility values, a set of actions, and a set of perceived environmental states.

Unlike other cognitive models, the Model of Expertise introduced here can be applied to biological experts (humans) and to artificial experts (cognitive systems and artificial intelligence). Because the Model of Expertise is based on abstract Expertise Level and Knowledge Level descriptions, implementation details of how an entity carries out a skill or implements a knowledge store is not specified. This leaves implementation details up to the entities themselves. Humans implement the skills and knowledge stores quite differently than computers do and different cognitive systems developed in the cog era will implement them differently from each other.

The Model of Expertise introduced here can be applied to other cognitive architectures and cognitive models as well as is demonstrated in the next section. Our hope is the Model of Expertise can serve as a common ground and common language facilitating comparison, contrast, and analysis of different systems, models, architectures, and designs.

This version of the Model of Expertise contains 12 fundamental skills (13 including the *collaborate* skill) and 14 knowledge stores. Future research may identify additional skills and additional knowledge stores and we invite collaborative efforts along these lines of thought.

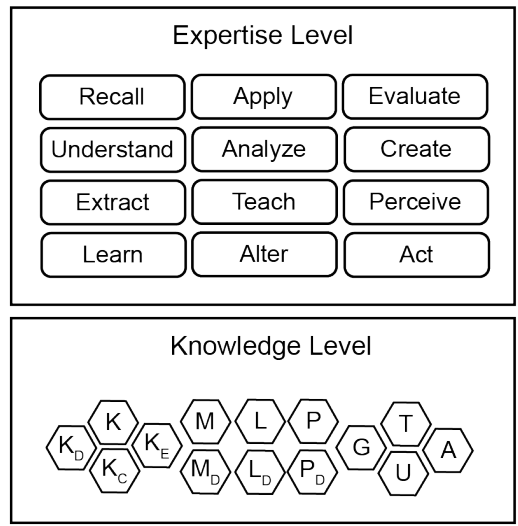

*Knowledge*
**K**      declarative knowledge statements
**$K_D$**   domain-specific knowledge
**$K_C$**   common-sense knowledge
**$K_E$**   episodic knowledge
**M/$M_D$** world models
**L/$L_D$** task models
**P/$P_D$** problem-solving models

**G**  goals to achieve
**U**  utility values
**T**  perceivable states
**A**  actions

*Skills*
**Perceive**   sense/interpret the environment
**Act**        perform action affecting environment
**Recall**     remember; store/retrieve knowledge
**Understand** classify, categorize, discuss, explain, identify
**Apply**      implement, solve, use knowledge
**Analyze**    compare, contrast, experiment
**Evaluate**   appraise, judge, value, critique
**Create**     design, construct, develop, synthesize
**Extract**    match/retrieve deep knowledge
**Learn**      modify existing knowledge
**Teach**      convey knowlege/skills to others
**Alter**      modify goals

**Fig. 8**. The Model of Expertise

### 3.5 Composite Processes/Activities

An expert certainly performs more actions than the skills identified in the Model of Expertise. It is important to note the skills listed in the Model of Expertise are fundamental in nature. Other, higher-level processes and activities are composites combining one or more fundamental skills. Examples are the *justify* activity and the *predict* process. Whenever a human or an artificial system arrives at a decision, it is common for someone to ask for justification as to why that decision was made. To justify a decision or action, the expert would exercise a combination of the *recall*, *analyze,* and *understand* skills. To *predict* an outcome, the expert would exercise a combination of *recall, analyze, evaluate*, and *apply* skills. Future research may very well identify additional fundamental skills. However, in doing so, care should be taken to identify skills which cannot be composed of combination of the fundamental skills in the model. Other examples of composite processes/activities include:

| | |
|---|---|
| Check me | Conjecture |
| Clarify | Conceptualize |
| Define | Debate |
| Emphasize | Exemplify |
| Explain how | Explain when |
| Explain where | Explain why |
| Expand the scope | Expound/Elucidate |
| Gather evidence of | Gather information on |
| Give me alternatives | Give me analogies |
| How do you feel about | Illustrate/Depict |
| Inspire me | Make a case for |
| Make a case against | Make me feel better about |
| Monitor and notify me | Motivate me |
| Narrow the scope | Organize |
| Predict | Prioritize |
| Show me | Show me associations |
| Simplify | Summarize |
| Theorize | Think differently than me |
| Visualize | What if |
| What is best for me | What is the cost of |
| What is most important | What is this least like |
| What is this most like | |

## 4      Applications

This section demonstrates ways to employ the Model of Expertise introduced in this paper. We certainly encourage others to apply the Model of Expertise in different ways in their own research.

### 4.1 The Soar Model of Expertise

Cognitive scientists have studied and modeled human cognition for decades. The most successful cognitive architecture to date, begun by pioneer Allen Newell and now led by John Laird, is the Soar architecture shown in Fig. 2. The Model of Expertise can be applied to and incorporated into the Soar architecture as shown in Fig. 9. The knowledge stores are located in long-term memory (***T*** and ***A*** are shown associated with the *perceive* and *act* functions) and are brought into short-term working memory by one or more of the skills. Reinforcement, semantic, and episodic learning continually updates the knowledge stores in long term memory. Higher-order processes defined in Soar are composite processes resulting from a combination of the fundamental skills in the Model of Expertise. Future research should document and analyze these processes and ground them to the skills and knowledge stores identified in the Model of Expertise introduced here.

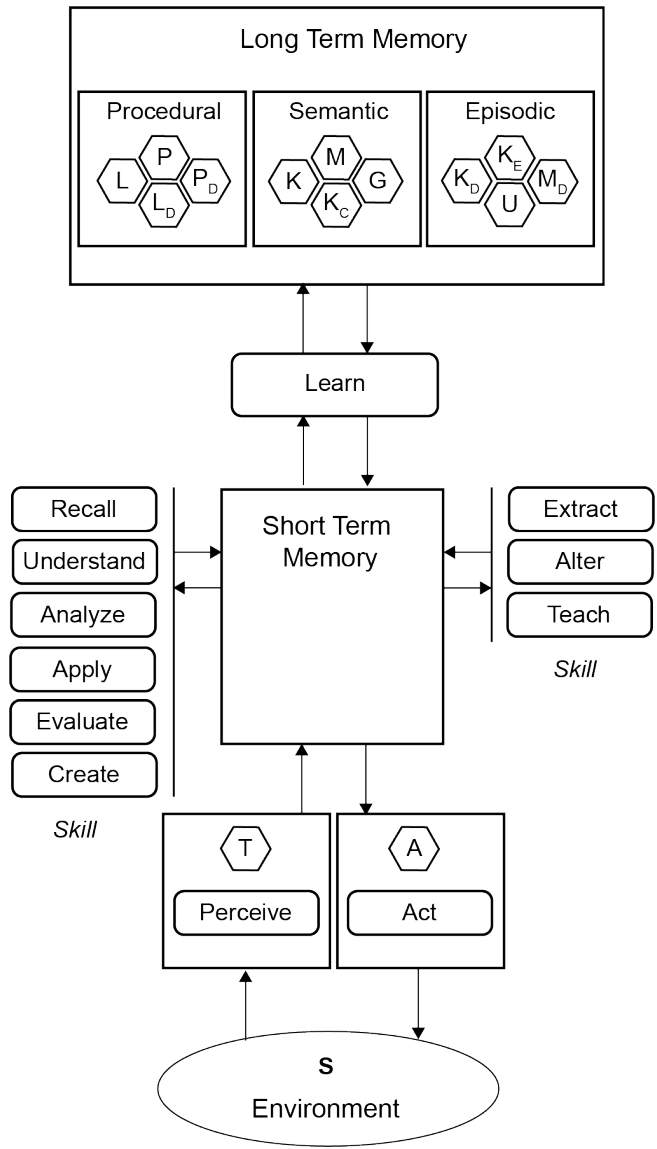

**Fig. 9**. Soar Model of Expertise

### 4.2 Cognitive Augmentation/Synthetic Expertise

We will soon be surrounded by artificial systems designed for the mass market capable of cognitive performance rivaling or exceeding a human expert in specific domains of discourse. Indeed, we see the beginning of this era now with voice-activated assistants and applications on our smartphones and other devices. John Kelly, Senior Vice President and Director of Research at IBM describes the coming revolution in cognitive augmentation as follows [29]:

> "The goal isn't to replace human thinking with machine thinking. Rather humans and machines will collaborate to produce better results—each bringing their own superior skills to the partnership."

The future lies in humans collaborating with artificial systems capable of high-level cognition. Engelbart was one of the first who thought of a human interacting with an artificial entity as a system [30]. While working together on a task, some processes are performed by the human (explicit-human processes), others are performed by artificial means (explicit-artifact processes), and others are performed by a combination of human and machine (composite processes). In the cognitive systems future, cognition will be a mixture of biological and artificial thinking. The human component in this ensemble will be said to have been cognitively augmented. We can represent cognitive augmentation using our Model of Expertise as shown in Fig. 10.

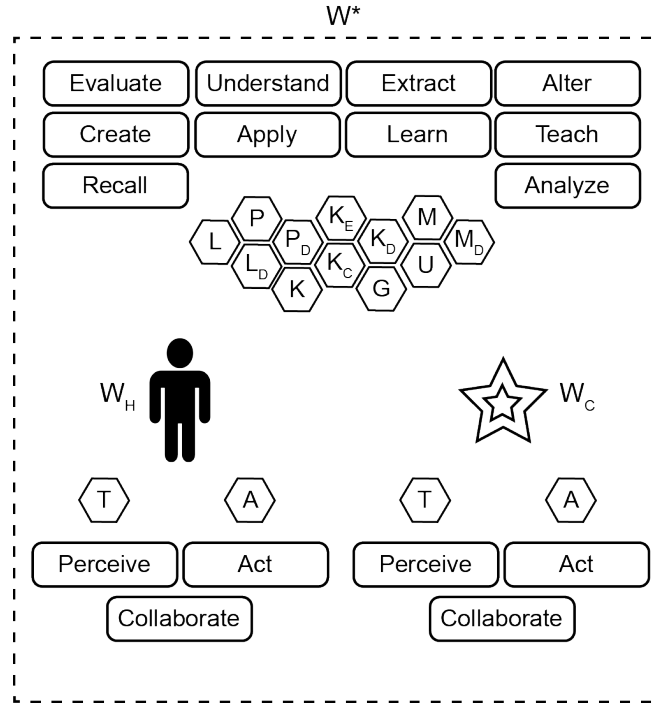

**Fig. 10**. Synthetic Expertise

The figure depicts a human working in collaboration with an artificial entity, a cognitive system, called a cog. As in Engelbart's framework, some of the skills are performed by the human and some are performed by the cog. In some cases, portions of a skill are performed by both the human and the cog. The human performs an amount of the *cognitive work* ($W_H$) and the cog performs an amount of the cognitive work ($W_C$). The cognitive work performed by the entire ensemble is $W^*$. The most important thing is all skills identified in the Model of Expertise are performed by the human/cog ensemble. It does not matter to the outside world whether or not a biological or artificial entity performs a skill.

Because the human and the artificial system are physically independent entities, we have drawn each with perceive and act skills and an additional skill, *collaborate*, has been added. This is necessary because in order to work together the human and the artificial entity must collaborate. In the figure, the knowledge stores are represented as not belonging solely to either entity. Physically, the human will have its own version of all knowledge stores in the formal model and the cog will have its own knowledge stores. However, logically, the knowledge of the ensemble will be a combination of the human and artificial knowledge sources. In fact, the entire cognitive performance of the human/cog ensemble is the emergent result of human/artificial collaboration and cognition. When the ensemble can achieve results exceeding most in the population it will have achieved the Gobet definition of expertise described earlier of achieving results superior to most. The ensemble will have achieved *synthetic expertise*.

An average human and non-expert in a field of study acting alone is able to perform to a certain level. The same human working in collaboration with a cog will be able to perform at a higher level, even to the level of an expert. Therefore, to the world outside of the human/cog ensemble, the human will appear to be *cognitively augmented*.

**4.3 Cognitive System Architectures**

The Model of Expertise introduced in this paper can be used to design future cognitive systems. An example is Lois, an artificial companion and caretaker for the elderly shown in Fig. 11.

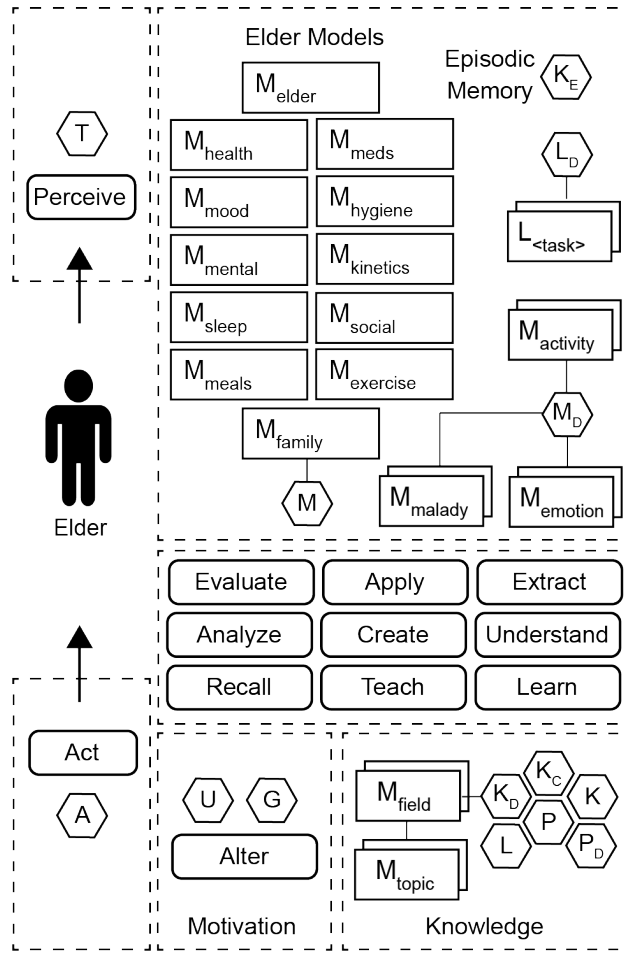

**Fig. 11**. Synthetic Elderly Companion (Lois)

Lois is composed of several different models allowing it to monitor the status and well-being of the elder. Episodic memory allows Lois to remember every interaction with the elder and use the knowledge to learn new knowledge, tasks, and activities. We invite researchers to develop more such models based on the Model of Expertise

# 4 Discussion

In the introduction, several recent and impressive success stories were discussed featuring an artificial system performing better than human experts in a particular domain. Using the Model of Expertise introduced in Fig. 8, we can now discuss these and other systems in new ways. The Model of Expertise identifies fundamental skills and types of knowledge required of an expert. How do real systems stack up against the model?

The systems discussed in the introduction, DeepBlue, Watson, AlphaGo, AlphaGo Zero, and AlphaZero, all perform some form of *perceive*, *analyze*, *evaluate*, and *act* skills. However, the scope of these functions is limited to the domain of discourse—the state of the gameboard and scoring the effect of a move. The problem here is the limited scope of the games themselves. Even though these systems have achieved expert-level performance, the scope of the expertise is small. Most systems in existence today share this characteristic of being narrowly focused. We call this *narrow expertise*. Systems able to learn general principles applicable to other domains is a very active area of research. This line of thought leads to another question. Should we expect an artificial expert to perform well in other domains? We do not necessarily expect this from human experts. For example, if an expert plumber is a terrible carpenter, we do not think less of him or her as a plumber.

No system discussed in the introduction can be said to *create*. If a Chess or Go program plays a sequence of moves never before played in the history of the game, did the system create it or just stumble upon it? This leads to an interesting argument involving understanding and this is a debate that has raged throughout the history of artificial intelligence research. Few would agree DeepBlue, Watson, AlphaGo, AlphaGo Zero, and AlphaZero actually understand the games they play and many would argue true creation cannot happen without understanding.

However, recently, systems have been developed to create faces (thispersondoesnotexist.com), music (www.aiva.ai), art (obviousart.com), and news stories (openai.com/blog/better-language-models). These systems do not understand what they are doing and their sole purpose is to create. Do these systems possess a true *create* skill? The images of the fake people created are indistinguishable from photographs of real people. Likewise, the music, art, and language are indistinguishable from that generated by humans and therein lies the problem. Recalling the Gobet definition of expertise, experts are expected to achieve results superior than most others. Therefore, an argument against these systems is the generated faces, music, portraits, and news briefs are not better, they are simply the same as those created by humans. This might pass the Turing test criterion involving artificial behavior a human cannot distinguish from human behavior, but the criterion for expertise is higher according to Gobet's definition. With experts, we are looking for exemplar behavior.

Unfortunately, whether not an object is an exemplar is highly subjective in most cases. However, we would not be surprised if, soon, an artificial system creates a hit song or an award-winning bit of poetry or prose. Even if systems achieve something like this, the question still remains: do systems understand what they have done? That brings us back to the above question. Strong AI proponents maintain systems must know and understand what they are doing to be considered intelligent whereas weak AI proponents require only results. Therefore, at best, all known systems at this time are examples of weak expertise.

None of the systems discussed so far exhibit the *teach* skill. The field of intelligent tutoring systems has been an active area of research for several decades and several tutoring systems exist, especially in mathematics. However, these systems are designed and built solely for the purpose of teaching specific knowledge/skills in a particular domain, not to be an expert in the domain. These systems are not expert mathematicians. Therefore, today we have systems achieving expertise but are not teachers and we have systems that teach but are not experts.

So far we have focused discussion on the skills in the Model of Expertise. What about the knowledge of an expert? All game-playing systems have some form of goals and utility value mechanism to execute strategy. However, a system's ability to *alter* and assess/*evaluate* new goals is limited to the game play context. There are only a few goals of interest in game play. For example, AlphaGo would never synthesize a goal of "acquire a ham sandwich." Artificial expertise has not yet been achieved in domains where complex goals and utility values are necessary. This again speaks to the narrow expertise and weak expertise nature of systems today as discussed earlier.

One of the great achievements of systems like AlphaGo, AlphaGo Zero, and AlphaZero has been the knowledge acquired via semi-supervised or unsupervised machine deep learning ($K_D$). Because of their narrowness, these systems do not learn generic knowledge like common-sense knowledge nor general knowledge ($K, K_C$). Again, no known system requiring extensive common-sense or generic knowledge has achieved expert-level performance.

Neither do these systems acquire episodic knowledge ($K_E$). For example, a human Go player would be able to recount a particularly interesting game against an opponent in which they learned an effective bit of strategy. A human would be able to tell you when the game was, who the opponent was, what they were feeling at the time, what was happening in the world about that time, etc. This is deep episodic memory—capturing the entire experience. Researchers are working on experience and context capture, common-sense learning, and general knowledge acquisition, but as of yet, these systems appear to be separate efforts no yet included into an artificial expert.

The degree of problem-solving knowledge ($L, L_D$) in these systems is difficult to assess without detailed analysis of how they work. One can certainly make the case these systems can learn and recall solutions and even strategies. But again, the narrowness of the game playing domain means these systems are not learning generic problem-solving knowledge ($L$) and any domain-specific problem-solving knowledge ($L_D$) is limited in scope and breadth.

Task knowledge is similarly restricted in game-playing systems. Since the set of actions, ($A$) is limited to game-related movement and placement of pieces, these systems do not have to perform an array of different kinds of tasks. Consider the difference between AlphaGo and an expert human carpenter. An expert knows how to perform hundreds if not thousands, of tasks related to carpentry (e.g. ways to make different kinds of joins), knows how to use dozens, if not hundreds, of different kinds of tools, and knows dozens of ways to apply paint, sealer, and varnish. The narrowness of existing systems limits task and problem-solving knowledge.

Game-playing systems certainly recognize situations and respond accordingly. For example, a Chess program will recognize its king is in check. It is unclear, but probably not the case, these systems use templates, frames, or other kinds of dynamic symbolic models. Medical diagnosis systems involve mostly pattern recognition and classification. For example, when detecting cancer in a radiograph, an artificial system will detect a pattern (part of its *perceive* and *understand* skillset) and then compare it (*match*, *analyze*, *evaluate*) to examples of known cancerous radiographs. If deploying an artificial neural network, the network's response to a stimulus (the radiograph) is compared to responses to known cancerous radiographs. Therefore, modeling seems to be built into the system itself rather than exist as a standalone knowledge store.

In the Model of Expertise, the extract skill represents the ability of an expert to *match* perceptions to stored knowledge and then retrieve a quantity of deep domain, procedural, and task knowledge ($P, P_D, L, L_D$). The systems discussed do not perform in this manner primarily because they do not seek to implement an entire expert (a doctor or game-playing person). Instead, today's systems should be considered to be low-level cogs better suited to be employed in a human/cog ensemble to achieve synthetic expertise.

## 5    Conclusion

A new abstract level, the Expertise Level, has been introduced to lie above Newell's Knowledge Level. Skills are the medium at the Expertise Level. An abstract Knowledge Level and Expertise-Level description of expertise has been introduced drawing from previous research in cognitive science, artificial intelligence, cognitive architectures, and cognitive augmentation. The Knowledge Level description describes the kinds of knowledge stores an expert must possess and the Expertise Level description describes the skills an expert must possess. Together, the Knowledge Level and Expertise Level description of expertise forms the Model of Expertise. The Model of Expertise can be used in all fields involving the study of intelligence and cognition including: cognitive science, cognitive architectures, artificial intelligence, cognitive systems, and cognitive augmentation. Specifically, we would like to see the Model of Expertise and the Expertise Level used to guide and assess future systems capable of artificial expertise and synthetic expertise, particularly in the cognitive systems area.

Integration of our model of expertise into the Soar cognitive architecture has been demonstrated. This could facilitate development of artificial expertise systems and exploration of human cognition based on the Soar architecture. Also demonstrated was using the Model of Expertise in the field of human cognitive augmentation to describe synthetic expertise whereby cognitive output of the ensemble is a mixture of biological and artificial cognition.

Finally, we discussed and an analyzed current game-playing systems using the Model of Expertise identifying them as examples of narrow expertise. Using the Model of Expertise, we may now distinguish narrow expertise from broad expertise and weak expertise from strong expertise by the degree to which systems utilize the knowledge and all skills identified at the Expertise Level of the Model of Expertise. Future cognitive systems will be developed for domains requiring extensive use of all skills and knowledge stores defined in the Model of Expertise.